\documentclass[conference]{IEEEtran}
\ifCLASSINFOpdf
\else
\fi

\renewcommand\IEEEkeywordsname{Keywords}

\IEEEoverridecommandlockouts
\usepackage{cite}

\usepackage{amsmath,amssymb,amsfonts}
\usepackage{amssymb}
\usepackage{breqn}
\usepackage{cleveref}
\usepackage[lined,ruled,linesnumbered,commentsnumbered]{algorithm2e}
\usepackage{booktabs}
\usepackage{cases}
\usepackage{graphicx}
\usepackage{multirow}
\usepackage{mathrsfs}
\usepackage{textcomp}
\usepackage{xcolor}
\usepackage{subfigure}
\usepackage{boxedminipage}

\def\BibTeX{{\rm B\kern-.05em{\sc i\kern-.025em b}\kern-.08em
    T\kern-.1667em\lower.7ex\hbox{E}\kern-.125emX}}

\begin{document}


\title{Online Bagging for Anytime Transfer Learning}

\bibliographystyle{IEEEtran}

\author{\IEEEauthorblockN{Guokun CHI, Min JIANG*, Xing GAO,\\
		 Weizhen HU, Shihui GUO*}
\IEEEauthorblockA{School of Informatics\\
Xiamen University\\
Xiamen, China 361005\\
minjiang@xmu.edu.cn, guoshihui@xmu.edu.cn}
\and
\IEEEauthorblockN{Kay Chen TAN}
\IEEEauthorblockA{Department of Computer Science\\
City University of Hong Kong\\
Hong Kong SAR, China}
\thanks{*Min JIANG and *Shihui GUO are the corresponding authors of this paper.}
}

\maketitle

\begin{abstract}

Transfer learning techniques have been widely used in the reality that it is difficult to obtain sufficient labeled data in the target domain, but a large amount of auxiliary data can be obtained in the relevant source domain. But most of the existing methods are based on offline data. In practical applications, it is often necessary to face online learning problems in which the data samples are achieved sequentially. In this paper, We are committed to applying the ensemble approach to solving the problem of online transfer learning so that it can be used in anytime setting. More specifically, we propose a novel online transfer learning framework, which applies the idea of online bagging methods to anytime transfer learning problems, and constructs strong classifiers through online iterations of the usefulness of multiple weak classifiers. Further, our algorithm also provides two extension schemes to reduce the impact of negative transfer.  Experiments on three real data sets show that the effectiveness of our proposed algorithms.

\end{abstract}


\section{INTRODUCTION}

In the field of data mining and machine learning, transfer learning \cite{weiss2016survey} has been widely studied as one of the important research topics. The classification performance of a target data set (target domain) that does not easily acquire sufficient labeled samples is improved by training one or more related auxiliary sample sets (referred to as source domains). Integrating the relevant source domain instance into the training model\cite{dai2007boosting}, or by mapping the training model of the source domain to the target model through domain adaptation\cite{pan2010domain}, can obtain effective knowledge transfer.

Most existing research on transfer learning \cite{dai2007boosting, pan2010domain, long2013transfer, wu2013cotransfer, jiang2015integration} focuses on the setting of offline (batch) learning, where the training data sets of the source domain and target domain are assumed to be given in advance. However, in real-world applications, training examples in many applications come in sequential order. If you want to get all the data at once, you often can not get it or pay a high price. Therefore, research on efficient online transfer learning algorithms that use only a few target examples is receiving more and more attention.

In the past ten years, much important research work on online transfer learning includes \cite{zhao2014online, wang2015online, yan2017online, chen2018heteotl}. To reduce the challenge of online transfer learning: there is no guarantee that the classification effect after the transfer will be improved, because incorrect source domain may lead to negative learning (negative transfer). Many researchers have adopted a combination of transfer learning technology and other related machine learning techniques to achieve maximum positive transfer and improve the final classification effect. The learning framework represented by the online transfer learning algorithm (OTL) \cite{zhao2014online} is currently the most popular. Boyu \textit{et al.} \cite{wang2015online} extended the transfer integration approach to the online version.

In this paper, we focus our research on online transfer learning in the context of homogeneous data. We propose a new algorithm OTBag. Applying the classic bagging algorithm to the problem setting of online transfer learning, the advantages of the bagging algorithm are fully utilized, and a more complex strong classifier model is constructed through a series of weak classifiers. To some extent, it overcomes the disadvantage that the PA \cite{crammer2006online} algorithm in OTL, which is limited to a single classifier model that cannot capture more complex hierarchies\cite{shi2017online}. At the same time, to prevent the negative transfer problem caused by transfer learning, two filtering strategies (SDMV, JDSMV) are proposed for the final model screening stage after weak classifier online training.

The rest of the paper is organized as follows: we introduce the knowledge of the relevant fields in section II. In Section III, we introduce our algorithm OTBag and two filtering strategies. We introduce the experimental researchs in Section IV and conclude in Section V.


\section{RELATED WORK}
\subsection{Online Transfer Learning}

Online transfer learning includes two important branches in the field of machine learning research, namely online learning and transfer learning.

Online learning is distinguished from many classic machine learning methods by its own good efficiency and scalability. It has been attracting attention for many years \cite{cesa2004generalization, zhao2011double, yang2010online, oza2005online}. Online learning focuses on solving some practical problems that training data is achieved in order. The online algorithms widely used by people can be divided into three categories: a. perceptron-based online learning algorithms; b. support vector machine (SVM-based) online algorithms; c. an online algorithm based on ensemble learning.

Transfer learning is to use knowledge from the source domains (as auxiliary knowledge) to improve the learning performance of the target domain \cite{weiss2016survey}. According to different learning settings, various transfer learning methods that have been proposed can be divided into three categories: inductive transfer learning\cite{niculescu2007inductive}, transductive transfer learning\cite{daume2006domain}, and unsupervised transfer learning\cite{samanta2013cross}. As a tool, transfer learning skill is widely used in other fields of machine learning, including anti-transfer in deep learning \cite{li2018learning} and robot motion generation task \cite{vyas2018neural, li2018learning}.

Online transfer learning is a combination and breakthrough of traditional online learning and transfer learning. Not only that, but many current research efforts have begun to combine online transfer learning with traditional learning algorithms, or to extend offline transfer learning algorithms to online versions to deal with real-world problems in online scenarios. The paper \cite{shan2018online} uses ensemble learning and active learning to build a stable online learning framework to solve the problem drift problem in the online data flow. Yan\textit{et al.} \cite{yan2017online} solves online heterogeneous transfer learning tasks by building a classifier by combining the weighted ensemble methods of offline and online decision making. In the paper \cite{wang2015online}, the traditional offline form of Boosting for Transfer Learning (TrAdaboost) is modified into an online transfer boosting method combined with the promotion method.

\subsection{Online Bagging}

 The online version of Bagging \cite{oza2005online} is an extension of the traditional offline bagging \cite{breiman1996bagging}. For offline bagging, the entire training bag is ready to be used, because, for each basic model, the sampling is performed by randomly attracting the entire training set. In bagging, each original training example can be repeated 0 times, 1 time, 2 times or more in each basic training set. The bootstrap training set for each basic model contains K copies of each original training example. Using this, Oza \textit{et al.} turned the problem into the following form,
\begin{equation}\label{eq:aei2}
    P\left( K=k \right) =\left( \begin{array}{c}
    	N\\
    	k\\
    \end{array} \right) \left( \frac{1}{N} \right) ^k\left( 1-\frac{1}{N} \right) ^{N-k}
\end{equation}
which represents the binomial distribution. By randomly extracting the entire training set, one instance at a time, and using Equation (1) to select K represents K times of resampling for the current instance. This can be equivalent to replace the N batch replacement sampling of the entire data set in the traditional batch bagging. Due to the unknown of the training sample N, the training samples are constantly coming, making the Equation (1) unavailable. However, the training samples in the online training application are coming, so we can assume $ N\rightarrow \infty $, and the binomial distribution will tend to be Poisson(1) distribution: $ P\left( K=k \right) =\frac{\exp \left( -1 \right)}{k!}$. At this time, the dependence on the total amount of samples N in the Equation (1) can be eliminated. For each new instance in the online training, $K\sim Poisson\left( 1 \right)$ is used to generate the number K of updates to the base classifier. The final classifier is the same as the batch Bagging and also uses the majority voting mechanism, $h\left( x \right) =arg\max _{y\in Y}\sum\limits_{m=1}^M{I\left( h_m\left( x \right) =y \right)}$.

Online bagging is a good approximation of batch bagging algorithm because their sampling methods produce an approximate bootstrap training set distribution, and when the training sets have similar distributions, their basic model learning algorithms will produce similar hypothesis spaces.

\section{PROPOSED ALGORITHM}

The proposed online transfer bagging (OTBag) algorithm is an extension of the online bagging algorithm. In \cite{kamishima2009trbagg}, the author first proposed to introduce transfer learning into the bagging algorithm under the batch data set and to improve the target training instances through a large number of source domain instances.
The problem, while the effective and diversified source data can reduce the target domain training error and improve the performance of the classifier. Our algorithm is inspired by this.

By using the knowledge of the source domain for the target domain, it is expected that the variance part of the error can be better reduced. However, in the face of the shortcomings of the transfer learning itself, there is still a negative transfer problem in the online transfer problem. Since the introduction of an instance in the source domain that is not related to the concept of the target domain will not only promote the construction of the target classifier but may lead to a worse final result. Ideally, we want to be able to identify those instances that are irrelevant during the training process, but this is not possible. But we can add a filtering strategy to the classifier, which will reduce the impact of negative transfer to a certain extent.

To better explain our ideas, we refer to the process of learning weak classifiers and the selection process of constructing final classifications as the stages of training and filtering. We will describe it in detail below. The source domain examples are represented as $ S_S=\left\{ \left( x_{1}^{S},y_{1}^{S} \right) ,...,\left( x_{N_S}^{S},y_{N_S}^{S} \right) \right\}$, and $ S_T=\left\{ \left( x_{1}^{T},y_{1}^{T} \right) ,...,\left( x_{N_T}^{T},y_{N_T}^{T} \right) \right\}$ is represented as a target domain. $N_S$ and $N_T$ represent the total number of source and target domain instances, respectively. And $S_S,S_T\in \mathbb{R}^d\times \left\{ 0,1 \right\} ,\ N_S\gg N_T\ $. In the training set, the labels of the source domain instance and the target instance are known, but target domain training examples is a small amount.

\textbf{\emph{1) Training Phase:}} In the training phase, the source training instance $S_S$ and the target training instance $S_T$ are integrated into the final training set $D=S_S\cup S_T$. The order of the samples is randomly scrambled, but the identity that identifies the instance from the target domain is retained. During the training process, for each new sample, it is judged that it belongs to the training domain of the source domain or the target domain. If it comes from the target domain, then it is used to train the $H=\left\{ h_1,h_2,...,h_M \right\}$ model, and train the target instance into the $F=\left\{f_1, f_2,...,f_M\right\}$ model, where M is the number of weak classifiers. For the source domain samples, only put them into model $H$ for training. See the $6-10$ lines in Algorithm 2 for details. At the same time, in the $\emph{H}$ and $\emph{F}$ models, it is necessary to record the correct classification of each new target instance, which is represented by $Acc_{h_i}\left( i=1,..., M \right)$ and $Acc_F$, respectively. Specifically, for the $H$ model, it is necessary to record whether each weak classifier $h_i$ is correctly classified for the target instance. And for the $F$ model, there is only one indicator, that is, the classification of the newly arrived target training instance by the final classifier constructed by the $F=\left\{f_1, f_2,...,f_M\right\}$ model through the majority voting mechanism. Algorithm 1 OTBag does not adopt a filtering strategy that reduces negative transfer, so it does not require an extra training $F$ model. Since the data is continuously coming in the form of online, the accuracy of the prediction of the target instance in the training set is also updated and recorded in real time.

\begin{algorithm}
  \caption{Online Transfer Bagging (OTBag)}
  \label{alg:rtm}
  \KwIn{$S_S, S_T, M$}
  \For{$n=1,2,...$}{
  Receive $\left(x_n, y_n \right)$ \;
  \For{$m=1,2,...,M$}{
  Let $k\sim Poisson\left( 1 \right)$\;
  Do $k$ time
  \mbox{}\par
    ~~~~Update the base learner $h_m$, using $\left(x_n,y_n\right)$\;
    }
  }
  \Return
      $\textbf{h}\left( x \right) =arg\max _{y\in Y}\sum\limits_{m=1}^M{I\left( h_m\left( x \right) =y \right)}$
\end{algorithm}

\textbf{\emph{2) Filtering Phase:}} At the end of the training phase, we can get the two models $H$, $F$ after training.
In the filtering phase, we need to make different strategic choices for $H=\left\{ h_1,h_2,...,h_M \right\}$ weak classifiers to more accurately predict the labeling of the target concept and effectively reduce the impact of negative transfer.
As with traditional batch and online bagging methods, it is most common to combine the final M by a majority voting strategy, as shown in our algorithm 1. But this strategy applies to situations where the source domain and the target domain are conceptually as similar as possible, i.e. the source domain instance can have a positive impact on the target instance. However, we are unable to confirm whether the given source instance is similar to the target instance concept, so we propose two other strategies that can preserve the advantages of ensemble learning for classification performance improvement while also reducing the impact of negative transfer.
\begin{algorithm}
  \caption{Simple Dominant Majority Voting (OTBag-SDMV)}
  \label{alg:rtm}
  \KwIn{$S_S, S_T, M$}
  Initialize $Acc_{h_m}=0, Acc_F=0$ for all $m\in \left\{ 1,...,M \right\}$\;
  \For{$n=1,2,...$}{
  Receive $\left(x_n, y_n \right)$ \;
  \For{$m=1,2,...,M$}{
  Let $k\sim Poisson\left( 1 \right)$\;
  Do $k$ time
  \mbox{}\par
    ~~Update the base learner $h_m$, using  $\left(x_n,y_n\right)$\;
     \uIf{$\left( x_n,y_n \right) \in S_T$}{
        Upadate the $f_m$, using$\left(x_n,y_n\right)$\;
        \lIf{$h_m\left( x_n\right)=y_n$}{
        $Acc_{h_m} = Acc_{h_m}+1$\;
        }
    }
    }
    $\textbf{F}\left( x \right) =arg\max _{y\in Y}\sum\limits_{m=1}^M{I\left( f_m\left( x \right) =y \right)}$\;
    \lIf{$\textbf{F}\left( x_n\right)=y_n$}{
        $Acc_F = Acc_F+1$\;
        }
  }
    \textbf{for} all $m\in \left\{ 1,...,M \right\}$ \\
    ~~~~\lIf{$Acc_{h_m}\ge Acc_F$}{
        $H^*\longleftarrow H^*\cup h_m$\;
        }
  \Return
      $\textbf{h}^*\left( x \right) =ara\max _{y\in Y}\sum_{h_i\in H^*}{I\left( h_i\left( x \right) =y \right)}$

\end{algorithm}

\textbf{\emph{2.1) Simple Dominant Majority Voting （SDMV):}} The classification accuracy rate of $Acc_{h_i}\left( i=1,..., M \right)$ maintained by $H$ is directly compared with the training set accuracy $Acc_F$. If a weak classifier $h_i$ has a worse empirical error than the $F$ model on the target training set $S_T$, the $h_i$ is rejected directly, regardless of its use to construct the final classifier. The M weak classifiers are compared with the $F$ model in turn. Finally, the dominant weak classifier set $H^*$ is left for constructing the final strong classifier. As with the traditional online bagging method, the majority voting is used for $H^*$ to construct the final model. See Algorithm 2 for details.

\textbf{\emph{2.2) Joint Double Subset Majority Voting （JDSMV):}} This method calculates the classification of the training target instances in each segment by dividing the online training learning process into multiple time segments. In this way, different weak classifier combinations can be selected. Finally, a subset of multiple weak classifiers can be formed. The online training process including the source domain and the target domain is divided into $\alpha$ phase. $\alpha$ is a specified hyperparameter, indicating the number of time segments. $\eta =\left( N_S+N_T \right) /\alpha$, $\eta$ represents the number of training samples included in each time segment. As shown in Algorithm 3, the models $H$ and $F$ are simultaneously trained (initialized) in the training set in the first time segment $seg_1$, starting from the second time slice $seg_2$, for the target domain training instance in the current time slice, the classification accuracy of all the weak classifiers $h_i$ in the $H$ model and the online classification accuracy of the $F$ model are separately recorded. The weak classifier index set $index_i$ with higher recording accuracy than the $F$ model, such as $index_2={1, 3, 5}$, indicated that the classification accuracy of the weak classifiers $h_1,h_3,h_5$ to the target training instance in the time segment $seg_2$ is higher than that of the $F$ model. All the time segments are completed in turn, that is, all training set instances are trained to form a total index set $\zeta=\left\{ index_2,index_3,...,index_{seg_{\alpha}} \right\}$. The strong classifier is constructed by forming a subset of weak classifiers corresponding to the indexes recorded in each time segment in $\zeta$, and adopting a majority voting strategy in this subset to obtain a final decision. Finally, the weak classifier corresponding to the index $index_i$ recorded under each time segment will constitute a subset, and a majority voting strategy is used in this subset to obtain a decision $\psi_{seg_i}\left( i=2,3,...,\alpha \right)$. At the same time, the $F$ model can obtain a decision $\psi_F$ by using the majority voting strategy. Finally, in the decision $\psi_{seg_i}$, $\psi_F$, a total of $\left({\alpha-1} \right)$ strong classifiers through the majority voting strategy to generate the final classification model.

Compared with the majority voting strategy used by the traditional bagging algorithm (which can be considered as a layer set), the joint majority vote of the double subset here can recombine the weak classifiers that dominate (biased target instances). Achieve further filtering of weak classifiers with negative effects.

\begin{algorithm}
  \caption{Join Double Subset Majority Voting (OTBag-JDSMV)}
  \label{alg:rtm}
  \KwIn{$S_S, S_T, M$}
  Initialize $\alpha, \eta =\left( N_S+N_T \right) /\alpha, \zeta=\left\{\right\}, Acc_{h_m,i}=0, Acc_{F,i}=0$ for all $m\in \left\{ 1,...,M \right\}$, $i\in \left\{2,...,\alpha \right\}$\;

  \For{$n=1,2,...$}{
  Receive $\left(x_n, y_n \right)$ \;
  \For{$m=1,2,...,M$}{
  Let $k\sim Poisson\left( 1 \right)$\;
  Do $k$ time
  \mbox{}\par
    ~~Update the base learner $h_m$, using $\left(x_n,y_n\right)$\;
     \uIf{$\left( x_n,y_n \right) \in S_T$}{
        Upadate the $f_m$, using $\left(x_n,y_n\right)$\;
        \lIf{$h_m\left( x_n\right)=y_n$}{
        $Acc_{h_m,i} = Acc_{h_m,i}+1$\;
        }
    }
    }
    $\textbf{F}\left( x \right) =arg\max _{y\in Y}\sum\limits_{m=1}^M{I\left( f_m\left( x \right) =y \right)}$ \;
    \lIf{$\textbf{F}\left( x_n\right)=y_n$}{
        $Acc_{F,i} = Acc_{F,i}+1$\;
        }
    \uIf{$n> \eta ~\textbf{and}~ n\% \eta=0$}{
    \lIf{$Acc_{h_m,i}\ge Acc_{F,i}$}{
        $index_i\leftarrow index_i\cup m, i=i+1$\;
    }
  }
  $H^{**}\gets$ \\
  $\quad \quad H^{**}\cup  H_i^* \left(majority\ voting\ on\ subset\ h_{index_i} \right)$

  }

  \Return
      $\textbf{h}^{**}\left( x \right) =ara\max _{y\in Y}\sum_{h_i\in H^{**}}{I\left( h_i\left( x \right) =y \right)}$

\end{algorithm}

\section{EXPERIMENTS}

In this section, the performance of our proposed algorithm will be verified on three data sets. In all experiments, a large number of source domain instances and a small number of target domain instances were directly integrated and trained in an online form. In order to reduce the random impact of the training sequence of the example on the final result, the final experimental result comes from the average data of 20 sets of randomly arranged training sets and test sets. We set the number of iterations $M$ and $\alpha$ in the comparison algorithm to 10. The training set and test set partition ratio in the experiment is 4:6.
\begin{table*}[htbp]
  \centering
  \caption{Accuracies (mean \% $\pm$ standard deviations) on Sentiment Analysis Data Sets}
    \begin{tabular}{|c|c|c|c|c|c|c|}
    \toprule
    \multicolumn{1}{|p{6.25em}|}{\textbf{           Algorithm Tasks }} & HomOTL-I & HomOTL-II & OTB   & \textbf{OTBag} & \textbf{OTBag-SDMV} & \textbf{OTBag-JDSMV} \\
    \midrule
    b $\rightarrow$ d & 53.27$\pm$0.02 & 51.58$\pm$0.02 & 61.73$\pm$0.01 & \textbf{76.00$\pm$0.01} & 57.80$\pm$0.00 & 75.67$\pm$0.00 \\
    \midrule
    b $\rightarrow$ e & 64.94$\pm$0.02 & 66.75$\pm$0.02 & 65.13$\pm$0.01 & 72.27$\pm$0.02 & 58.33$\pm$0.00 & \textbf{73.80$\pm$0.01} \\
    \midrule
    b $\rightarrow$ k & 64.63$\pm$0.01 & 65.92$\pm$0.01 & 69.67$\pm$0.01 & 75.40$\pm$0.01 & 58.67$\pm$0.01 & \textbf{76.93$\pm$0.00} \\
    \midrule
    d $\rightarrow$ b & 48.19$\pm$0.02 & 48.67$\pm$0.04 & 62.80$\pm$0.01 & 71.20$\pm$0.01 & 57.80$\pm$0.01 & \textbf{71.67$\pm$0.01} \\
    \midrule
    d $\rightarrow$ e & 64.73$\pm$0.01 & 66.65$\pm$0.02 & 66.07$\pm$0.01 & 76.13$\pm$0.02 & 58.67$\pm$0.01 & \textbf{77.53$\pm$0.01} \\
    \midrule
    d $\rightarrow$ k & 53.04$\pm$0.02 & 52.13$\pm$0.02 & 65.60$\pm$0.02 & 73.93$\pm$0.02 & 58.47$\pm$0.01 & \textbf{75.93$\pm$0.01} \\
    \midrule
    e $\rightarrow$ b & 49.83$\pm$0.01 & 48.75$\pm$0.03 & 66.67$\pm$0.01 & \textbf{70.47$\pm$0.01} & 57.73$\pm$0.00 & 70.13$\pm$0.01 \\
    \midrule
    e $\rightarrow$ d & 51.12$\pm$0.02 & 49.31$\pm$0.02 & 60.60$\pm$0.01 & 71.73$\pm$0.02 & 58.33$\pm$0.01 & \textbf{72.00$\pm$0.01} \\
    \midrule
    e $\rightarrow$ k & 48.02$\pm$0.02 & 48.67$\pm$0.04 & 70.33$\pm$0.02 & 76.80$\pm$0.03 & 58.33$\pm$0.00 & \textbf{77.27$\pm$0.00} \\
    \midrule
    k $\rightarrow$ b & 65.04$\pm$0.02 & 67.75$\pm$0.02 & 62.80$\pm$0.01 & \textbf{71.93$\pm$0.01} & 58.80$\pm$0.00 & 71.87$\pm$0.01 \\
    \midrule
    k $\rightarrow$ d & 59.21$\pm$0.01 & 59.21$\pm$0.01 & 65.93$\pm$0.01 & \textbf{72.40$\pm$0.02} & 58.47$\pm$0.00 & 71.67$\pm$0.01 \\
    \midrule
    k $\rightarrow$ e & 70.73$\pm$0.01 & 73.44$\pm$0.02 & 61.20$\pm$0.01 & 74.67$\pm$0.01 & 59.27$\pm$0.01 & \textbf{77.87$\pm$0.01} \\
    \bottomrule
    \end{tabular}%
  \label{tab:addlabel}%
\end{table*}%

\subsection{Data Sets}

\textbf{Sentiment analysis data set} \cite{chen2012marginalized} is composed of Amazon users' evaluation of the four types of products: books, DVDs, electronics, and kitchen $(b, d, e, k)$. Each review contains a user's rating (0-5 stars), review title, product name, reviewer name, location, date, and comment content. For ratings with scores greater than 3 stars, positive instances, and instances with ratings below 3 stars are marked as negative instances, and other comments will be discarded because their ratings are not clear. The other preprocessing for the data set is the same as in \cite{chen2012marginalized}, and the feature dimension of the sample is 400, and each domain contains 2000 positive/negative sample sets. We selected 50\% as the total number of training sets and test sets used in the experiment. We use the symbol $ b\rightarrow d $ to generate the source domain from books (b) and the target domain from DVDs (d). Each domain is randomly selected as the source domain and the target domain, so 12 transfer learning tasks can be generated.

\textbf{Object recognition data set} \cite{gong2012geodesic} uses 2 object image data sets to our algorithm, namely: Amazon and Caltech-256 \cite{griffin2007caltech}. There are 10 public classes in each dataset. Similar to the article \cite{cui2014flowing}, we extract the SURF feature from the image set, encode the image using the 800-bin histogram, and finally normalize the feature and z-scored. We treat each data set as a domain, from which one domain is selected as the source domain and one as the target domain. The symbol $A\rightarrow C $ is used to represent the source domain generated from Amazon (A) and the target domain is generated from Caltech-256 (C). We select the two adjacent classes into two classification problems in order, there are five groups, BACKPACK vs TOURING-BIKE, CALCULATOR vs HEAD-PHONES, 	KEYBOARD vs LAPTOP-101, MONITOR vs MOUSE, COFFEE MUG vs VIDEO-PROJECTOR. For each class in the dataset, 60\% was selected as the test set and the rest as the training set.

\textbf{Mixed data set:} In order to more clearly verify the performance of the negative transfer data, the above two data sets are mixed to form a “third data set”. For example, the partial data of Amazon in object recognition and the books data of sentiment analysis (here intercepted 400-dimensional features) are mixed into a new source domain dataset, and the DVDs dataset of sentiment analysis is used as the target domain and is expressed as $mix\_b\rightarrow d $. Mixing the object recognition sample into the source domain formed by the sentiment analysis instance will be very different from the target domain, and can not directly learn by the method of inductive transfer, and even a significant negative transfer phenomenon will occur.

\subsection{Experimental Results }

In the experiment, we focused on the classification effect of the algorithms OTBag, OTBag-SDMV, OTBag-JDSMV on three data sets. Compare the three algorithms proposed by us with the current most popular online transfer algorithms OTB \cite{wang2015online}, HomOTL-I, HomOTL-II \cite{zhao2014online}. On the one hand, the classification performance of the algorithm on the data set similar to the target concept in the first two source domains is compared, and on the other hand, the verification algorithm responds to the negative transfer effect under the mixed data set.

We can see from Table I that among the 12 transfer tasks under the sentiment analysis data set, the performance of the OTBag and OTBag-JDSMV algorithms is the best, and the accuracy between them is not much different. This is due to the fact that the forward source instance extends the diversity of the training samples, and the majority of the voting strategies of the final weak classifiers maintain this positive impact. Among them, the proposed OTBag algorithm has a relatively better improvement in accuracy than the current OTB algorithm and HomOTL-I, HomeOTL-II algorithm. We attribute the reason to the fact that the strong classifier built by multiple weak classifiers has better performance.

Compared with the direct majority voting of OTBag, the OTBag-JDSMV algorithm adopts a dual joint voting mechanism to construct a subset of weak classifiers according to the selection strategy, so that the final strong classifier is more robust. However, the limitation of this strategy is only to re-establish the combination of weak classifiers. The weak classifier itself has not been modified more, so the final effect is not much different from the OTBag algorithm.
It is worth noting that another strategy we proposed, OTBag-SDMV, is not outstanding in classification performance, and is only a little better than HomOTL-I/II in some tasks. The reason is that few weak classifiers have better classification effects on target training instances than the $F$ model. Here we propose that the purpose of this strategy is more to hope that it can perform better under the data set that cannot directly perform the inductive transfer, that is, the impact of "negative transfer" (explained in the experiment below).
\begin{table*}[htbp]
  \centering
  \caption{Accuracies (mean \% $\pm$ standard deviations) on Object Recognition Sets}
    \begin{tabular}{|c|c|c|c|c|c|c|}
    \toprule
    \multicolumn{1}{|c|}{\multirow{2}[4]{*}{\textcolor[rgb]{ .149,  .149,  .149}{\textbf{      
    				   Algorithm \& Tasks }}}} & \multicolumn{6}{c|}{BACKPACK vs TOURING-BIKE} \\
\cmidrule{2-7}          & \textcolor[rgb]{ .149,  .149,  .149}{HomOTL-I} & \textcolor[rgb]{ .149,  .149,  .149}{HomOTL-II} & \textcolor[rgb]{ .149,  .149,  .149}{OTB} & \textcolor[rgb]{ .149,  .149,  .149}{\textbf{OTBag}} & \textbf{OTBag-SDMV} & \textbf{OTBag-JDSMV} \\
    \midrule
    \textcolor[rgb]{ .149,  .149,  .149}{A $\rightarrow$ C} & \textcolor[rgb]{ .149,  .149,  .149}{\textbf{80.09$\pm$0.03}} & \textcolor[rgb]{ .149,  .149,  .149}{78.75$\pm$0.02} & \textcolor[rgb]{ .149,  .149,  .149}{61.09$\pm$0.15} & \textcolor[rgb]{ .149,  .149,  .149}{77.38$\pm$0.19} & \textcolor[rgb]{ .149,  .149,  .149}{58.80$\pm$0.00} & \textcolor[rgb]{ .149,  .149,  .149}{78.14$\pm$0.01} \\
    \midrule
    \textcolor[rgb]{ .149,  .149,  .149}{C $\rightarrow$ A} & \textcolor[rgb]{ .149,  .149,  .149}{91.47$\pm$0.03} & \textcolor[rgb]{ .149,  .149,  .149}{89.38$\pm$0.04 } & \textcolor[rgb]{ .149,  .149,  .149}{75.08$\pm$0.25} & \textcolor[rgb]{ .149,  .149,  .149}{93.11$\pm$0.14} & \textcolor[rgb]{ .149,  .149,  .149}{68.53$\pm$0.01} & \textcolor[rgb]{ .149,  .149,  .149}{\textbf{93.28$\pm$0.01}} \\
    \midrule
          & \multicolumn{6}{c|}{CALCULATOR vs HEAD-PHONES} \\
    \midrule
    \textcolor[rgb]{ .149,  .149,  .149}{A $\rightarrow$ C} & 69.64$\pm$0.06   & 76.48$\pm$0.03  & 62.40$\pm$0.21 & \textcolor[rgb]{ .149,  .149,  .149}{\textbf{86.71$\pm$0.22}} & \textcolor[rgb]{ .149,  .149,  .149}{57.87$\pm$0.01} & \textcolor[rgb]{ .149,  .149,  .149}{86.23$\pm$0.01} \\
    \midrule
    \textcolor[rgb]{ .149,  .149,  .149}{C $\rightarrow$ A} & 67.98$\pm$0.08   & 75.60$\pm$0.02  & 69.78$\pm$0.12 & \textcolor[rgb]{ .149,  .149,  .149}{\textbf{94.81$\pm$0.13}} & \textcolor[rgb]{ .149,  .149,  .149}{68.73$\pm$0.01} & \textcolor[rgb]{ .149,  .149,  .149}{94.52$\pm$0.01} \\
    \midrule
          & \multicolumn{6}{c|}{KEYBOARD vs LAPTOP-101} \\
    \midrule
    \textcolor[rgb]{ .149,  .149,  .149}{A $\rightarrow$ C} & 74.50$\pm$0.04 & 73.14$\pm$0.04  & 64.30$\pm$0.30 & \textcolor[rgb]{ .149,  .149,  .149}{74.09$\pm$0.36} & \textcolor[rgb]{ .149,  .149,  .149}{58.93$\pm$0.01} & \textcolor[rgb]{ .149,  .149,  .149}{\textbf{75.17$\pm$0.01}} \\
    \midrule
    \textcolor[rgb]{ .149,  .149,  .149}{C $\rightarrow$ A} & 76.67$\pm$0.05  & 76.20$\pm$0.03  & 54.21$\pm$0.25 & \textcolor[rgb]{ .149,  .149,  .149}{81.00$\pm$0.39} & \textcolor[rgb]{ .149,  .149,  .149}{58.33$\pm$0.01} & \textcolor[rgb]{ .149,  .149,  .149}{\textbf{81.14$\pm$0.01}} \\
    \midrule
          & \multicolumn{6}{c|}{MONITOR vs MOUSE} \\
    \midrule
    \textcolor[rgb]{ .149,  .149,  .149}{A $\rightarrow$ C} & 60.83$\pm$0.02 & 68.93$\pm$0.02  & 67.55$\pm$0.31 & \textcolor[rgb]{ .149,  .149,  .149}{\textbf{80.00$\pm$0.39}} & \textcolor[rgb]{ .149,  .149,  .149}{58.33$\pm$0.01} & \textcolor[rgb]{ .149,  .149,  .149}{79.75$\pm$0.01} \\
    \midrule
    \textcolor[rgb]{ .149,  .149,  .149}{C $\rightarrow$ A} & 65.85$\pm$0.05 & 63.70$\pm$0.06 & 63.02$\pm$0.28 & \textcolor[rgb]{ .149,  .149,  .149}{\textbf{93.09$\pm$0.44}} & \textcolor[rgb]{ .149,  .149,  .149}{58.87$\pm$0.01} & \textcolor[rgb]{ .149,  .149,  .149}{92.37$\pm$0.01} \\
    \midrule
          & \multicolumn{6}{c|}{COFFEEMUG vs VIDEO-PROJECTOR} \\
    \midrule
    \textcolor[rgb]{ .149,  .149,  .149}{A $\rightarrow$ C} & 56.30$\pm$0.05 & 66.97$\pm$0.02 & 67.55$\pm$0.31 & \textcolor[rgb]{ .149,  .149,  .149}{\textbf{79.22$\pm$0.38}} & \textcolor[rgb]{ .149,  .149,  .149}{58.13$\pm$0.00} & \textcolor[rgb]{ .149,  .149,  .149}{78.60$\pm$0.01} \\
    \midrule
    \textcolor[rgb]{ .149,  .149,  .149}{C $\rightarrow$ A} & 60.46$\pm$0.04 & 60.81$\pm$0.03 & 63.02$\pm$0.28 & \textcolor[rgb]{ .149,  .149,  .149}{81.49$\pm$0.39} & \textcolor[rgb]{ .149,  .149,  .149}{59.00$\pm$0.01} & \textcolor[rgb]{ .149,  .149,  .149}{\textbf{81.79$\pm$0.01}} \\
    \bottomrule
    \end{tabular}%
  \label{tab:addlabel}%
\end{table*}%

In Table 2, the accuracy performance of each algorithm under the object recognition data set is shown. Among the 5 groups of 10 tasks, 9 of them have the optimal performance belonging to OTBag and OTBag-JDSMV algorithms. As discussed above, ensemble learning has a richer meaning representation under the current data set than HomOTL-I/II using a single PA algorithm. For the OTB algorithm, which also uses the ensemble learning boosting method, its overall performance is better than the HomOTL-I/II algorithm. As for the reason why the OTB algorithm does not perform as well as the OTBag algorithm proposed by us, we attribute it to the problem of premature convergence of the source domain of the TrAdaboost algorithm \cite{al2011adaptive}. It is worth noting that we found that the accuracy of the OTBag-JDSMV algorithm is not much different from that of the OTBag algorithm, but its standard deviation is much smaller than the OTBag. This also verifies that our majority voting strategy for the dual subset of OTBag-JDSMV can improve the stability of OTBag calculations while ensuring accuracy.

\begin{table*}[htbp]
  \centering
  \caption{Accuracies (mean \% $\pm$ standard deviations) on Mixed Data Sets}
    \begin{tabular}{|c|c|c|c|c|c|c|}
    \toprule
    \multicolumn{1}{|p{6.25em}|}{\textbf{           Algorithm Tasks }} & HomOTL-I & HomOTL-II & OTB   & \textbf{OTBag} & \textbf{OTBag-SDMV} & \textbf{OTBag-JDSMV} \\
    \midrule
    mix\_b $\rightarrow$ d & 56.21$\pm$0.02 & 55.29$\pm$0.02 & 51.33$\pm$0.00 & 51.33$\pm$0.25 & \textbf{58.40$\pm$0.01} & 58.27$\pm$0.01 \\
    \midrule
    mix\_b $\rightarrow$ e & 48.62$\pm$0.02 & 48.42$\pm$0.02 & 52.67$\pm$0.00 & 47.33$\pm$0.23 & \textbf{59.07$\pm$0.01} & 58.67$\pm$0.01 \\
    \midrule
    mix\_b $\rightarrow$ k & 50.81$\pm$0.01 & 50.15$\pm$0.02 & 49.40$\pm$0.01 & 51.00$\pm$0.25 & 58.20$\pm$0.01 & \textbf{63.87$\pm$0.01} \\
    \midrule
    mix\_d $\rightarrow$ e & 50.08$\pm$0.02 & 49.04$\pm$0.02 & 50.00$\pm$0.00 & 50.00$\pm$0.24 & 58.80$\pm$0.00 & \textbf{61.80$\pm$0.01} \\
    \midrule
    mix\_e $\rightarrow$ k & 46.71$\pm$0.02 & 48.02$\pm$0.03 & 50.60$\pm$0.03 & 53.00$\pm$0.26 & 58.33$\pm$0.00 & \textbf{66.47$\pm$0.01} \\
    \midrule
    mix\_k $\rightarrow$ d & 57.04$\pm$0.02 & 56.81$\pm$0.02 & 50.40$\pm$0.02 & 52.00$\pm$0.25 & 59.13$\pm$0.01 & \textbf{60.40$\pm$0.00} \\
    \bottomrule
    \end{tabular}%
  \label{tab:addlabel}%
\end{table*}%

In the mixed data set, our main purpose is to demonstrate the performance of our algorithm for the effects of negative transfer. As shown in Table 3, among the six tasks under mixed data, both the SDMV and JDSMV strategies make our proposed OTBag perform better than the benchmark algorithm. Among them, OTBag-JDSMV is more prominent and dominates among the four tasks. The performance of the original OTBag algorithm that does not use the filtered negative transfer strategy is similar to that of the baseline algorithm. At this time, the source instance samples play a more negative role. For the two filtering strategies we have extended, they can show the effect of reducing the negative transfer and improving the classification accuracy under the current data set.

\section{CONCLUSION}
In this paper, we proposed the online transfer learning framework OTBag based on bagging. The algorithm has better classification accuracy than the popular single source domain online transfer learning method. At the same time, we have extended the two strategies (OTBag-SDMV / JDSMV) for the filtering phase of OTBag for the impact of smaller negative transfer. Among them, the Joint double subset majority voting (OTBag-JDSMV) strategy has outstanding performance in the above three data sets.

Our algorithm has a good performance on all three real data sets, but the solution is limited to the binary classification. In the future, we will introduce the multi-category problem setting, so that our algorithm can better match the real problem of multi-label in reality. At the same time, the idea of ​​introducing multiple source domains will be considered to further improve the performance of the algorithm.

\section*{Acknowledgment}
This work was supported by the National Natural Science Foundation of China (Grant No.61673328) and Shenzhen Scientific Research and Development Funding Program (Grant No. JCYJ20180307123637294).

\bibliography{mybibtex}

\end{document}